\documentclass[10pt,journal,compsoc]{IEEEtran}
\ifCLASSOPTIONcompsoc
  \usepackage[nocompress]{cite}
\else
  \usepackage{cite}
\fi
\ifCLASSINFOpdf
   \usepackage[pdftex]{graphicx}
\else
   \usepackage[dvips]{graphicx}
\fi
\usepackage{amsmath}

\usepackage{ragged2e}
\usepackage{amssymb}
\usepackage{mathtools}
\usepackage{xspace}
\usepackage{stix}
\usepackage{caption}
\usepackage{float}

\newcommand{\head}[1]{\noindent\textbf{#1}}

\DeclarePairedDelimiterX{\infdivx}[2]{(}{)}{%
  #1\;\delimsize\|\;#2%
}

\DeclareMathOperator*{\argmin}{argmin}

\makeatletter
\DeclareRobustCommand\onedot{\futurelet\@let@token\@onedot}
\def\@onedot{\ifx\@let@token.\else.\null\fi\xspace}

\def\eg{\emph{e.g}\onedot} 
\def\ie{\emph{i.e}\onedot} 
 
\def\etc{\emph{etc}\onedot}

\makeatother

\begin{document}
%
\title{3D Meta Point Signature: Learning to Learn 3D Point Signature for 3D Dense Shape Correspondence}
%
%
%
%

\author{Hao Huang,
        Lingjing Wang,
        Xiang Li,
        Yi Fang$^\ast$
\IEEEcompsocitemizethanks{\IEEEcompsocthanksitem Hao Huang is with the Department of Computer Science and Engineering, NYU, Brooklyn, NY, USA. Xiang Li, Lingjing Wang and Yi Fang are with the Department of Electrical and Computer Engineering, NYU Abu Dhabi, Abu Dhabi, UAE.\protect\\
E-mail: \{hh1811, lw1474, xl1845, yfang\}@nyu.edu
\IEEEcompsocthanksitem $\ast$ Corresponding author: Yi Fang (yfang@nyu.edu)}
}

%
%

\markboth{}%
{Shell \MakeLowercase{\textit{et al.}}: Bare Demo of IEEEtran.cls for Computer Society Journals}
%



\IEEEtitleabstractindextext{%
\begin{abstract}
\justifying{Point signature, a representation describing the structural neighborhood of a point in 3D shapes, can be applied to establish correspondences between points in 3D shapes. Conventional methods apply a weight-sharing network, \eg, any kind of graph neural networks, across all neighborhoods to directly generate point signatures and gain the generalization ability by extensive training over a large amount of training samples from scratch. However, these methods lack the flexibility in rapidly adapting to unseen neighborhood structures and thus generalizes poorly on new point sets. In this paper, we propose a novel meta-learning based 3D point signature model, named 3D \textbf{me}ta \textbf{p}oint \textbf{s}ignature (MEPS) network, that is capable of learning robust point signatures in 3D shapes. By regarding each point signature learning process as a task, our method obtains an optimized model over the best performance on the distribution of all tasks, generating reliable signatures for new tasks, \ie, signatures of unseen point neighborhoods. Specifically, the MEPS consists of two modules: a base signature learner and a meta signature learner. During training, the base-learner is trained to perform specific signature learning tasks. In the meantime, the meta-learner is trained to update the base-learner with optimal parameters. During testing, the meta-learner that is learned with the distribution of all tasks can adaptively change parameters of the base-learner, accommodating to unseen local neighborhoods. We evaluate the MEPS model on two datasets, \eg, FAUST and TOSCA, for dense 3D shape correspondence. Experimental results demonstrate that our method not only gains significant improvements over the baseline model and achieves state-of-the-art results, but also is capable of handling unseen 3D shapes.}
\end{abstract}

\begin{IEEEkeywords}
Point Signature, Shape Correspondence, Meta Learning, Graph Convolution.
\end{IEEEkeywords}}

\maketitle

\IEEEdisplaynontitleabstractindextext

%
\IEEEpeerreviewmaketitle

\begin{figure*}[!htb]
\centering
  \includegraphics[width=1\linewidth]{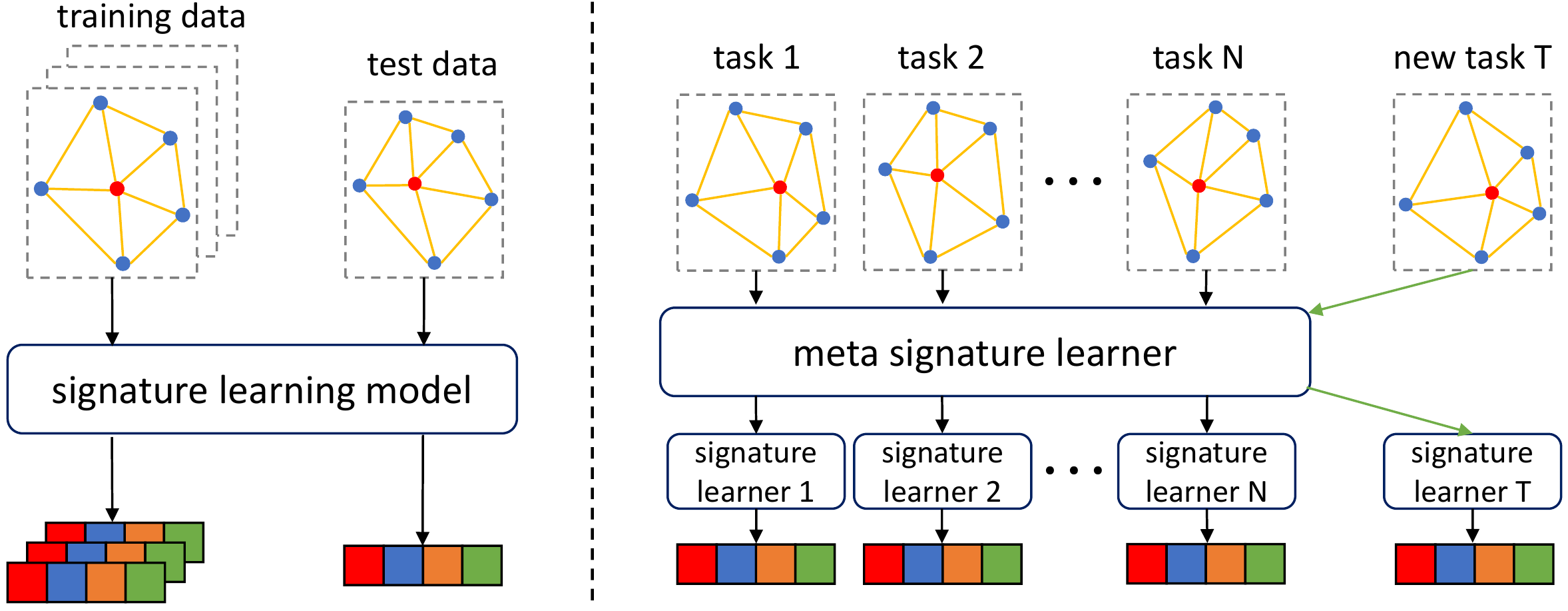}
  \caption{(Left) Conventional point signature learning process. (Right) Our 3D meta point signature (MEPS) network. Colored bar represents generated point signatures.}
  \label{fig:teaser}
\end{figure*}

\IEEEraisesectionheading{\section{Introduction}\label{sec:introduction}}
\IEEEPARstart{P}{oint} signature is defined as a representation that describes the structural neighborhood of a point in 3D shapes~\cite{chua1997point}, and this representation captures relations between neighboring points with the center point. Point signature can be applied to establish dense shape correspondences due to the similarity of signatures of corresponding points. Conventional methods generating point signatures often consist of two steps: 1) they first construct a neighborhood graph representing the local structure of each center point based on k-nearest neighbors, 2) and then apply one kind of graph neural networks (GNNs) to aggregate neighborhood point features. The aggregated feature is viewed as the signature of the center point within a local neighborhood patch. A typical network design scheme is to apply weight-sharing learnable layers across all points (and edges, in some cases) to compute per-point features. The learning process of such conventional methods is depicted in the left part of Figure~\ref{fig:teaser}. Specifically, a large amount of training samples are fed into a signature learning model, \ie, a graph neural network, to generate point signatures which are then used for subsequent tasks. After training, the optimized model whose parameters are fixed is evaluated on testing samples. However, such a network design scheme and the point signature learning process have two limitations. Firstly, it may fail to generate discriminative signatures for points whose neighborhoods are of subtle structural changes. A general formulation of a graph neural network can be decomposed into an ``update'' function and an ``aggregated'' function. The ``update'' function takes input as features of the center point and a neighboring point, producing a pair-wise feature, and the ``aggregation'' function aggregates all pair-wise features around the center point as its signature. As parameters of the ``update'' function are shared across all neighborhoods, they cannot adaptively change to exclusively describe each neighborhood's local structures in a fine-grained manner. Secondly, as parameters of the ``update'' function are optimized from scratch for all training samples and fixed after training, they cannot rapidly adapt to new testing samples, and thus cannot generalize well to produce reliable point signatures for points in unseen neighborhoods.

To address the limitations mentioned above, we propose to develop a novel meta-learning based 3D point signature model, dubbed as 3D \textbf{me}ta \textbf{p}oint \textbf{s}ignature (MEPS) network, which is capable of dynamically capturing neighborhood features including unseen ones and thus generating robust point signatures. The core idea of the proposed method is illustrated in the right part of Figure~\ref{fig:teaser}. Here we treat the process of learning point signature in each neighborhood as a task. Instead of directly mapping each center point along with its neighboring points into a point signature by a weight-sharing network model, we introduce a meta-learner to generate a group of models, denoted as base-learners, that achieve optimal performance for each corresponding task. Compared with conventional methods, the meta-learner in our MEPS network learns the distribution of all tasks and generates base-learner models that are dynamically tailored for unseen tasks based on the task distribution. Each specific point signature learning process is performed by a different base-learner, while the meta-learner learns to ``learn point signature''. The advantages of the MEPS network has two folds: fast generalization ability and fine-grained structure modeling ability. The base-learner can fast adapt to new point signature learning tasks as its parameters are updated adaptively by a meta-leaner which is optimized over the distribution of all tasks, rather than updated by directly trained over all local neighborhood samples from scratch and then fixed after training. Since point relations in different neighborhoods are depicted by different base-learners instead of a shared one, subtle structural changes can be captured, as each meta-learner's parameters are optimized based on its corresponding signature learning task.

Our main contributions are summarized as follows. First, we propose a meta-learning based 3D point signature learning method to learn the relations between neighboring points to generate signatures for center points. This method can rapidly adapt to new local structures and handle unseen neighborhoods. Note that the proposed method is generic enough and can be easily accommodated to any existing network operated on graphs. Second, we provide a theoretical proof justifying the necessity of the meta-learning process as well as how our learning method works better than conventional ones. Third, we evaluate the proposed method on a 3D vision dataset, \ie, FAUST~\cite{bogo2014faust} and TOSCA~\cite{bronstein2008numerical} datasets for 3D dense shape correspondence estimation, and achieve state-of-the-art performance by obtaining a significant improvement over the baseline model. The experiment results also validate that the proposed method is capable of handling unseen 3D shapes. Note that the proposed method can be further modified and adapted to 2D image correspondence~\cite{rocco2017convolutional,kim2018recurrent,chen2019arbicon} by treating 2D pixels as 3D points.

\section{Related Work}
\head{Deep learning on 3D data.} PointNet~\cite{qi2017pointnet} combines a multi-layer perceptron and a symmetric function to process the unordered 3D point sets. Instead of processing each point individually, later work~\cite{qi2017pointnet++,wang2019dynamic} propose to stack multiple PointNets hierarchically to aggregate neighborhood information and capture local area structures. Abstract away from PointNet used as basic building blocks, other methods~\cite{verma2018feastnet,li2018pointcnn,lan2019modeling,xu2018spidercnn} devise various convolution kernels or learnable modules to learn neighborhood features. SPLATNet~\cite{su2018splatnet} extends the idea of SPLAT~\cite{adams2010fast} from 2D domain to 3D domain; SONet~\cite{li2018so} incorporates the self-organization mechanism to ensure the invariance to point order. The seminal work~\cite{masci2015geodesic} introduces deep learning of local features on 3D shapes for correspondence and retrieval. DCP~\cite{wang2019deep} adopts Transformer~\cite{vaswani2017attention} to establish correspondences between different point clouds. PC-Net~\cite{li2019pc} utilizes an encoder to encode shape features and drifts points in a template to match two shapes using the encoded features. During the learning process for 3D data, representations of local structures, \ie, point signatures, are critical for high-level abstractions~\cite{chua1997point}. For instance, the corresponding two points on two shapes should have similar point signatures. In this work, we exclusively focus on the learning process of point signature for shape correspondence estimation.

\head{Graph-based models.} Graph neural network is one of the typical networks that operate on irregular data, including 3D meshes or point clouds. GNNs can be generally divided into two categories based on the domain where they are defined: spatial domain and spectral domain. In the first category, the work~\cite{niepert2016learning} extracts locally connected regions and transforms them into a canonical form processed by a conventional convolution neural network. EdgeConv~\cite{wang2019dynamic} explicitly extracts features for graph edges. TC-Net~\cite{li2020topology} introduces topology-associated edge weights to represent topological structure for each point. Spatial filtering introduced in~\cite{such2017robust} defines vertex filtering operated on vertices of a graph. In the second category, the core idea is convolution operation in spatial domain is equivalent to multiplication operation in spectral domain according to Convolution Theorem~\cite{mallat1999wavelet}. Methods falling into this category usually define Laplacian of graph~\cite{bruna2013spectral,defferrard2016convolutional,kostrikov2018surface}, and then transform graphs from spatial domain to spectral domain. Diffusion-convolution is introduced in~\cite{atwood2016diffusion} and applied on each node to determine its local neighborhood. The work~\cite{kipf2016semi} constructs spectral graph convolutions by using the first-order of Chebyshev polynomials. The method proposed in this paper falls into the first category, and we consider a point's neighborhood defined in spatial domain.

\head{Hypernetworks or dynamic networks.} Contrary to conventional deep learning where a given task is solved by training a model from scratch iterating over massive training samples, meta-learning aims to improve the model's fast-learning ability, given the experience of multiple learning tasks~\cite{hospedales2020meta}. Model-based methods in meta-learning rapidly update model's parameters by either its internal architecture~\cite{sukhbaatar2015end,santoro2016meta} or controlled by another meta-learner model~\cite{munkhdalai2017meta}. The applications of these methods include language modeling~\cite{ha2016hypernetworks}, image few-shot learning~\cite{munkhdalai2017meta}, and 3D shape representation~\cite{littwin2019deep,mitchell2020higher-order}. In this paper, we propose to formulate the point signature learning process under meta-learning framework and develop a robust learning method that can rapidly adapt to unseen point neighborhoods in new 3D shapes.

\section{Our Method}
In this section, we first revisit conventional methods learning point signatures using graph neural networks and a baseline model instantiating such networks~\cite{verma2018feastnet}. Then, we present our meta-learning based method in detail. Finally, we provide a theoretical analysis of our method and compare it with conventional methods.

\subsection{Conventional Point Signature Learning}
A graph built on a 3D point set is defined as a tuple $\mathcal{G}=(V, E)$ with $N^v$ vertices and $N^e$ edges. The $V=\{v_i\}_{i=1:N^v}$ is the set of vertices comprising of all points, where $v_i$ is vertex attribute. The $E=\{(e_{ik}, v_i, v_k)\}_{k\in\mathcal{N}(i)}$ is the set of edges, where $e_{ik}$ represents edge attribute and is associated with vertex $v_i$ and vertex $v_k$. The formulation of a graph neural network is expressed as~\cite{fey2019fast}:
\begin{equation}
\label{eq:gnn}
    v_i^\prime = h_\psi(v_i, \underset{k\in \mathcal{N}(i)}\dottedsquare f_\theta(v_i, v_k, e_{ik}))\enspace,
\end{equation}
where $\dottedsquare$ denotes a differentiable and permutation invariant ``aggregation'' function, \eg, mean, sum or max; $f_\theta$ and $h_\psi$ are differentiable ``update'' functions which are often parameterized as multi-layer perceptrons (MLPs). The learned feature $v^\prime_i$ is treated as the point signature for vertex $v_i$ and it captures the local structural information of neighborhood $\mathcal{N}(i)$ by modeling relations between all neighboring vertices $\{v_k\}_{k\in \mathcal{N}(i)}$ and $v_i$. 

A concrete instantiation of Eq.~\ref{eq:gnn} in FeaStNet~\cite{verma2018feastnet} is defined in Eq.~\ref{eq:feast_gnn} and illustrated in the left part of Figure~\ref{fig:pipeline} (For conciseness, $M$ is set to 1 in the figure.):
\begin{equation}
    v_i^\prime = \mathbf{b} + \frac{1}{\lvert \mathcal{N}_i \rvert}\sum_{k \in \mathcal{N}(i)}\big(\sum_{m=1}^M q_m(v_i, v_k)\mathbf{W}_m v_k\big)\enspace,
\label{eq:feast_gnn}
\end{equation}
where $\mathbf{W}_m$ and $\mathbf{b}$ represent weights and biases, respectively. The function $q_m$ models the relation between a pair of neighboring vertices ($v_i, v_k$) and is defined as a softmax over a linear transformation:
\begin{equation}
    q_m(v_i, v_k) = \exp(\mathbf{u}_m^\top v_i + \mathbf{v}_m^\top v_k + c_m)\enspace,
\label{eq:feast_q}
\end{equation}
where $\mathbf{u}_m$, $\mathbf{v}_m$ and $c_m$ are learnable parameters, and $M$ represents the total number of weights. The output of $q_m(v_i,v_k)$ is a scalar $r_{i,k}$ for pair-wise relation which is determined by the local structure of neighborhood $\mathcal{N}(i)$. Correspondingly, the learned $r_{i,k}$ should contain neighborhood information of $\mathcal{N}(i)$. The point signature for $v_i$ in this instantiation case is defined as a weighted-sum over all linear-transformed neighboring vertex features:
\begin{equation}
\text{ps}(v_i) = \sum_{k \in \mathcal{N}(i)}r_{i,k}\mathbf{W}v_k\enspace,
\label{eq:ps}
\end{equation}
where $\text{ps}(\cdot)$ stands for ``point signature'' for the input vertex.

Note that all parameters in the model are shared across all vertices, and point signatures are directly estimated by training the model over a large volume of training samples. A relational inductive bias~\cite{battaglia2018relational}, \ie, translation invariance, introduced by the weight-sharing structure impairs its ability to generate discriminative point signatures for distinct neighborhoods. Specifically, the resultant value of $q_m$ only depends on the attributes of two neighboring vertices $v_i$ and $v_k$, without considering the overall structural information of the neighborhood in which they locate. Another critical issue is that parameters of $q_m$ are optimized from scratch using training data and then fixed during testing, and cannot thus fast adapt to unseen neighborhoods to generate reliable signatures. To handle potentially large amounts of unseen neighborhood structures in a new 3D shape, we aim to design a model that generalizes well on unseen neighborhoods based on experience over the seen neighborhoods, rather than a model that merely gains optimal performance on the seen neighborhoods.

\subsection{Learning to Learn 3D Point Signature}
\begin{figure*}[!htb]
\centering
  \includegraphics[width=1\linewidth]{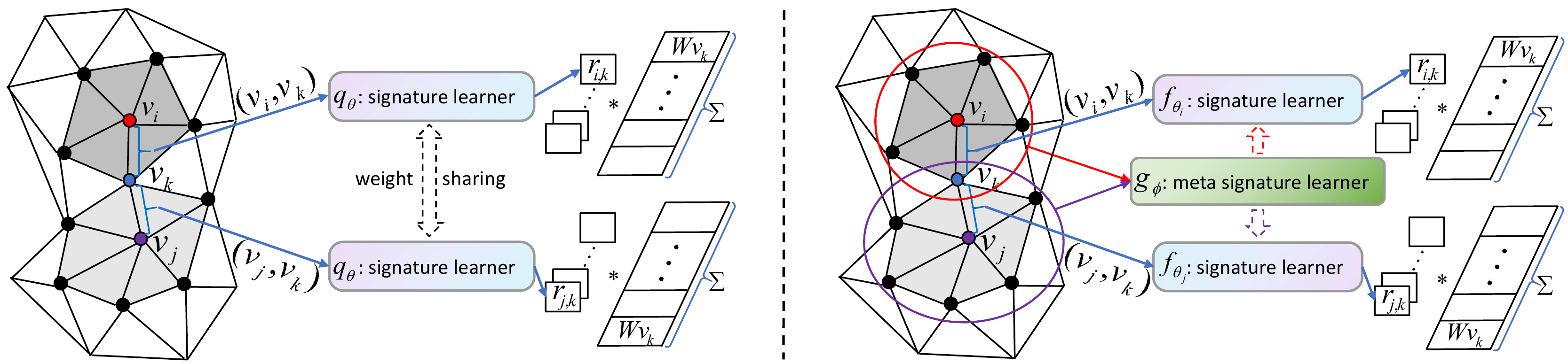}
  \caption{On the left, a weight-sharing signature learner network $q_\theta$ is applied to compute the relation value (\ie, $r_{i,k}$ or $r_{j,k}$) between a vertex (blue) and its central vertex (red or purple), which is then used to weight the convolutional feature of that vertex  ($Wv_k$) to add as the point signature for the central vertex (\ie, $v_i$ or $v_j$). In contrast, on the right, our proposed meta-signature learner $g_\phi$ firstly takes input a local neighbor (\ie, all vertices within the red circle or the purple circle) for a specific central vertex (\ie, red or purple) to dynamically updates the parameters of the base signature learner network (\ie, $f_{\theta_i}$ or $f_{\theta_J}$). The base signature leaner then performs vertex relation calculation to generate point signature for the central vertex.}
  \label{fig:pipeline}
\end{figure*}

To solve the issues mentioned above, we propose a novel meta-learning based 3D point signature learning model capable of learning robust point signatures in 3D shapes. In our model, the learning process of point signature for each neighborhood is treated as a separate task $\mathcal{T}$, and all vertices within a neighborhood compose a small dataset $\mathcal{D}$. Instead of applying a single weight-sharing network to directly map a vertex along with its neighbors into a signature as shown in the left part of Figure~\ref{fig:pipeline}, we devise a two-level network, named 3D \textbf{me}ta \textbf{p}oint \textbf{s}ignature (MEPS) network, to train over all tasks for signature generation in a two-stage style. Specifically, the MEPS consists of a meta-learner network $g_\phi$ and a base-learner network $f_\theta$, as shown in the right part of Figure~\ref{fig:pipeline}.
Given a neighborhood $\mathbf{p}=\{v_i\} \cup \{v_k \vert k \in \mathcal{N}(i)\}$, the meta-learner $g_\phi$ learns to generate a model, \ie, parameters of the base-learner $f_\theta$, that obtains the optimal performance on the task of signature generation for $\mathbf{p}$ in the first stage. The base-learner $f_\theta$ performs the specific task by estimating the relation between each neighboring vertex $v_j$ with the center $v_i$ to get the signature for $v_i$ in the second stage. These two stages are formally given by the following equations:
\begin{align}
    \theta_p &= g_\phi(\mathbf{p};\phi) \label{eq:meta_param}\enspace, \\
    r_{i,k} &= f_{\theta_p}(v_i, v_k;\theta_p)\enspace,
\end{align}
where $r_{i,j}$ represents the estimated relation between $v_i$ and $v_j$. Inside meta-learner $g_\phi$ in the first stage, a local feature embedding describing the overall structure of $\mathbf{p}$ is first extracted and then projected to the weight $\theta_p$. The model can be decomposed into two levels of neighborhood representation. The meta-learner $g_\phi$ is a mapping from input neighborhood $\mathbf{p}$ to the parameters $\theta_p$ which encode the structural information of $\mathbf{p}$, and the base-learner $f_{\theta_p}$ regresses each neighboring vertex $v_j$ to a value encoding its pair-wise relation with the center $v_i$. The concrete network architectures of $g_\phi$ and $f_\theta$ can be any form. Based on our proposed method discussed above, we re-formulate Eq.~\ref{eq:feast_gnn} as:
\begin{equation}
    v_i^\prime = \mathbf{b} + \frac{1}{\lvert \mathcal{N}_i \rvert}\sum_{k \in \mathcal{N}(i)}\big(\sum_{m=1}^M f^m_\theta(v_i, v_k)\mathbf{W}_m v_k\big)\enspace,
\label{eq:meta_gnn}
\end{equation}
where the point signature for the central vertex $v_i$ is defined as weighted-sum over transformed neighboring vertex features as in Eq.~\ref{eq:ps}. Despite that we adopt FeaStNet as a concrete instantiation of Eq.~\ref{eq:gnn} for illustration, our method is generic enough to be applied to any graph neural network built on Eq.~\ref{eq:gnn} with slight modification.

Note that even though the base-learner $f^m_\theta$ adopts the same architecture as $q_m$, they are still essentially distinct. In Eq.~\ref{eq:gnn}, parameters of $q_m$ are trained from scratch and shared across all neighborhoods $\mathcal{N}(i)$. On the contrary, parameters of $f_\theta$ in Eq.~\ref{eq:meta_gnn} are updated through $g_\phi$ by training over all signature learning tasks. The distribution of all tasks is denoted as $p(\mathcal{D})$. We expect $f_\theta$ to be capable of well adapting or generalizing to new tasks, \ie, unseen neighborhoods in new 3D shapes. Instead of keeping parameters of $f_\theta$ fixed after training, we adaptively update its parameters according to different tasks by employing a meta-leaner $g_\phi$ which learns over a variety of tasks and is optimized for the best performance on the distribution $p(\mathcal{D})$ of all tasks, including unseen ones. The optimal model parameters for $g_\phi$ are:
\begin{equation}
    \phi^\ast = \argmin_\phi \mathbb{E}_{\mathcal{D}\sim p(\mathcal{D})} [\mathcal{L}_\theta(\mathcal{D})]\enspace,
\end{equation}
where $\mathcal{L}_\theta(\mathcal{D})$ is task-related loss. Note that what $g_\phi$ eventually learns is the adapted model $f_\theta$ that learns to generate point signature, which is a process of ``learning to learn 3D point signature''.

\subsection{Effect of Meta-Learner}
We provide a theoretical proof in sketch that a shared-weight base-learner $f_\theta$ whose parameters are not updated by a meta-learner $g_\phi$ fails to incorporate the overall neighborhood information in modeling relations of neighboring vertices. 

As shown in Figure~\ref{fig:pipeline}, the vertex $v_k$ locates within two neighborhoods $\mathbf{p}_i=\{v_i\} \cup \{v_k \vert k \in \mathcal{N}(i)\}$ and $\mathbf{p}_j=\{v_j\} \cup \{v_k \vert k \in \mathcal{N}(j)\}$. Suppose $f_\theta$ is parameterized as an multi-layer perceptron (MLP), we remove $g_\phi$ and optimize parameters of $f_\theta$ across all neighborhoods in an input 3D shape. If we consider the pre-activation of the first layer of $f_\theta(v_i, v_k)$ and $f_\theta(v_j, v_k)$ can be expressed as:
\begin{align}
    \mathbf{a}^i_k &=
    \big[\hat{\mathbf{W}}:\tilde{\mathbf{W}}\big]\big[v_i:v_k\big]^\top + \mathbf{b} =  \hat{\mathbf{W}}v_i + \tilde{\mathbf{W}}v_k + \mathbf{b}\enspace, \\
    \mathbf{a}^j_k &= \big[\hat{\mathbf{W}}:\tilde{\mathbf{W}}\big]\big[v_j:v_k\big]^\top + \mathbf{b} =  \hat{\mathbf{W}}v_j + \tilde{\mathbf{W}}v_k + \mathbf{b}\enspace,
\end{align}
where $\big[\hat{\mathbf{W}}:\tilde{\mathbf{W}}\big]$ and $\mathbf{b}$ are parameters of this layer, $[:]$ denotes concatenation. As $\hat{\mathbf{W}}$, $\tilde{\mathbf{W}}$ and $\mathbf{b}$ share the same values when applied to pairs $(v_i, v_k)$ and $(v_j, v_k)$, the difference of pre-activation is:
\begin{equation}
\lVert \mathbf{a}_k^i - \mathbf{a}_k^j \rVert = \lVert \hat{\mathbf{W}}(v_i - v_j) \rVert \enspace,
\label{eq:act_diff}
\end{equation}
where $\lVert \cdot \rVert$ denotes a norm operator, \eg, $\ell_1$ or $\ell_2$ norm. Notice that the difference of pre-activation only depends on the central vertices of the two neighborhoods without considering neighboring vertices, including $v_k$ within the neighborhood. However, suppose the neighborhood $\mathcal{N}(i)$ locates on a flat plane while the neighborhood $\mathcal{N}(j)$ locates on a curved surface, ideally, the difference between $\mathbf{a}^i_k$ and $\mathbf{a}^i_k$ should reflect the geometric distinctness of the two neighborhoods, rather than just the difference between two central vertices. On the contrary, when $\theta_{p_i}$ and $\theta_{p_j}$ are dynamically generated by $g_\phi$ based on $\mathbf{p}_i$ and $\mathbf{p}_j$, they are different models and exclusively adapted to $\mathbf{p}_i$ and $\mathbf{p}_j$. Therefore, a term containing the information of $v_k$ will appear on the right-hand side of Eq.~\ref{eq:act_diff}.


\section{Experiments}
As the quality of point signatures directly determines the accuracy of point correspondence estimation, we evaluate our proposed method to generate point signature for 3D dense point correspondence on two 3D shape datasets.

\subsection{Dataset}
\head{FAUST.} The FAUST dataset~\cite{bogo2014faust} is a set of real human scanned meshes with ground-truth correspondences. It consists of 100 watertight meshes corresponding to 10 shapes in 10 different poses. Each mesh contains 6,890 vertices. We follow the same experimental setup in~\cite{boscaini2016learning,masci2015geodesic,monti2017geometric,verma2018feastnet} and formulate the shape correspondence problem between a reference shape and any other shape as a classification problem in which each vertex is annotated with one of 6,890 labels. We use the first 80 meshes as training data and the remaining 20 meshes of 10 poses of two unseen shapes for evaluation. The first shape in the first pose is used as the reference shape.

\head{TOSCA.} The TOSCA dataset~\cite{bronstein2008numerical} consists of high-resolution three-dimensional non-rigid shapes in a variety of poses by deforming template meshes. The database contains a total of 80 objects covering 8 categories. In the dataset, objects within the same category have the same number of faces and vertices numbered in a compatible way. To evaluate the performance of our model on non-human shapes for unseen classes, we conduct experiments by 1). training on cat shapes and testing on dog shapes; 2). training on horse shapes and testing on centaur shapes. The cat and dog sets contain 11 and 9 deformable shapes respectively; and the horse and centaur sets contain 8 and 6 deformable shapes respectively.

\subsection{Experiment Setup}
We adopt the same setting for both FAUST and TOSCA datasets. Concretely, the graph for each mesh in FAUST and TOSCA datasets is constructed using kNN algorithm by connecting k-nearest neighbors in which $k$ is set to 20. The input is raw 3D XYZ vertex coordinates for both datasets. We follow the network architecture of~\cite{verma2018feastnet} which consists of a sequence of linear layers and graph convolutions: Lin16+Conv32+Conv64+Conv128+Lin256+Lin6890. The numbers represent the output channels of each layer. Note that we only adopt the single-scale architecture with the number of filters $M=9$ (see Eq.~\ref{eq:feast_gnn}), rather than the multi-scale U-Net~\cite{ronneberger2015u} structure in~\cite{verma2018feastnet}. The key difference is that we replace the $q_m$ layer in Eq.~\ref{eq:feast_gnn} with our two-level networks. The base-learner network structure consists of two-layer linear transformations: (Lin32-Lin9) $\times$ 3 for each graph convolution. We apply PointNet~\cite{qi2017pointnet} as our meta-learner network to extract local patch features and generate the parameters of the base-learner network based on extracted local features. The feature size of each local patch is set to 1,024. We do not apply spatial/feature transformation, \ie, known as T-Net, in PointNet. Our model is trained using cross-entropy classification loss. We set the initial learning rate to 0.01 and weight decay to $10^{-4}$. We train the model with the batch size of 6 for a maximum of 6,000 epochs. The learning rate is decayed every 480 epochs with a factor of 0.5 and is clipped at $10^{-4}$.

\subsection{Results on FAUST}
\begin{figure}[!h]
\centering
\includegraphics[width=1\linewidth]{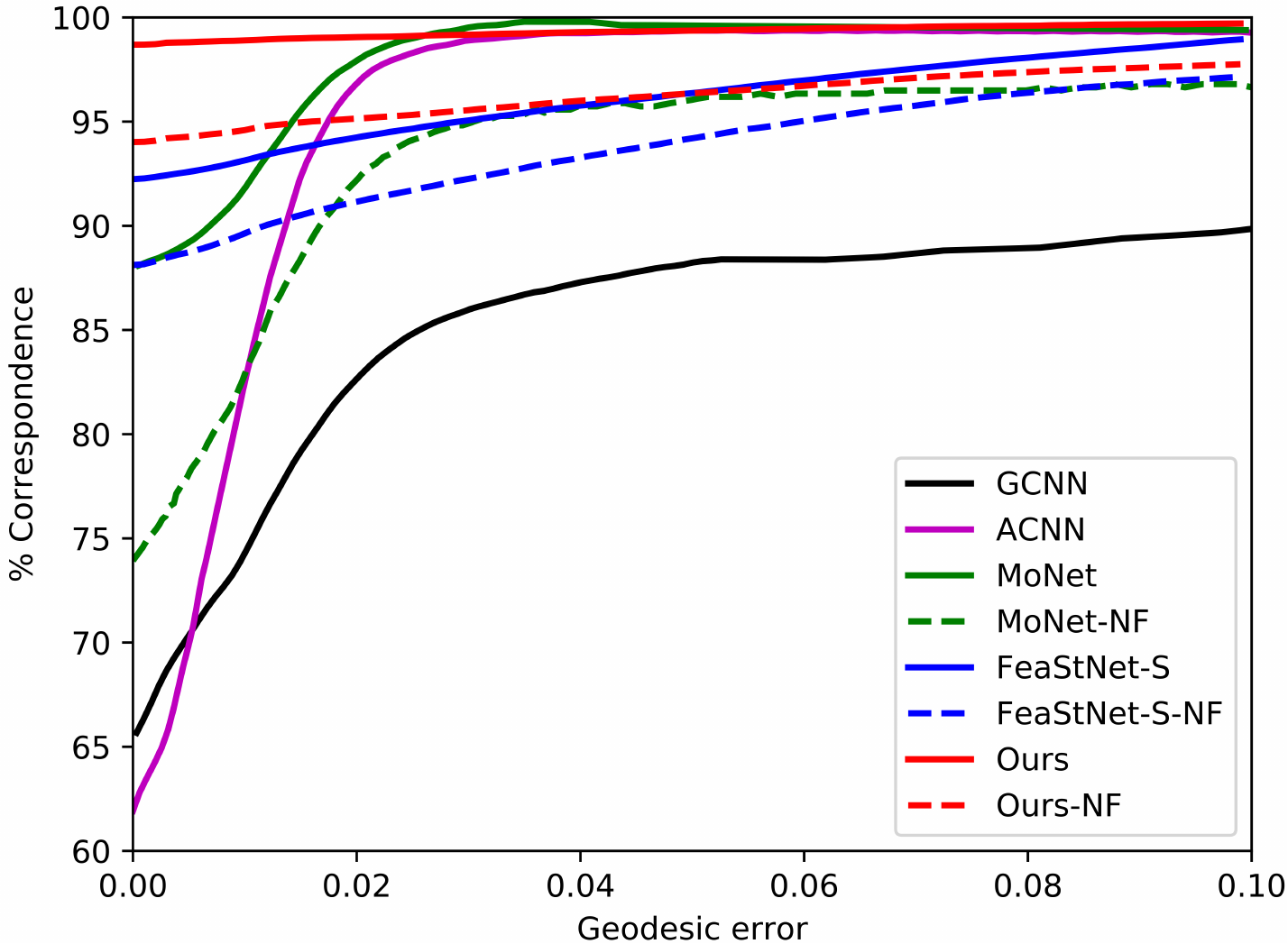}
\caption{Comparison of fraction of geodesic shape correspondence errors within a certain distance with state-of-the-art approaches. The ``-NF'' denotes results without refinement}
\label{fig:geo_dist}
\end{figure}

In Table~\ref{tab:faust_result}, we compare the correspondence accuracy with zero geodesic error obtained with our method using the single-scale architecture to state-of-the-art methods. For a fair comparison, we evaluate the quality of the correspondences directly predicted by our model without refinement, and with the refinement as post-processing using the algorithm proposed in~\cite{vestner2017product} which was also used by MoNet~\cite{monti2017geometric} and FeaStNet~\cite{verma2018feastnet}. Note that our single-scale model significantly outperforms the single-scale model in FeaStNet-S by 5.9\% without refinement. By applying refinement post-processing, our single-scale model achieves the same performance of 98.7\% with the multi-scale U-Net model in FeaStNet (not listed in Table~\ref{tab:faust_result}) and surpasses all other methods. It demonstrates the high-quality point signatures generated by our two-level meta-learning architecture through modeling relations between neighboring and central vertex. It also shows that our model can rapidly generalize on unseen tasks as the simpler model achieves the same results with the more complex one. In Figure~\ref{fig:geo_dist}, we plot the percentage of correspondences within a given geodesic distance from the ground-truth on the reference shape using Princeton benchmark protocol~\cite{kim2011blended}. Our method performs better than other methods under different geodesic error thresholds.

\begin{table}[!tb]
\caption{Comparison of correspondence accuracy on FAUST dataset of our model (single-scale) and the state-of-the-art approaches. Note that ``-S'' denotes single-scale architecture.}
\label{tab:faust_result}
\centering
\normalsize
\begin{tabular}{lcc}
\hline\hline
\textbf{Method}          & \textbf{Input} & \textbf{Accuracy} \\ \hline
Logistic Regression     & SHOT           & 39.9\%            \\
PointNet~\cite{qi2017pointnet}     & SHOT           & 49.7\%            \\
GCNN~\cite{masci2015geodesic}, w/o refinement     & SHOT           & 42.3\%            \\
GCNN~\cite{masci2015geodesic}, w/ refinement      & SHOT           & 65.4\%            \\
ACNN~\cite{boscaini2016learning}, w/o refinement     & SHOT           & 60.6\%            \\
ACNN~\cite{boscaini2016learning}, w/ refinement~\cite{ovsjanikov2012functional}      & SHOT           & 62.4\%            \\
MoNet~\cite{monti2017geometric}, w/o refinement    & SHOT           & 73.8\%            \\
MoNet~\cite{monti2017geometric}, w/ refinement~\cite{vestner2017product}     & SHOT           & 88.2\%            \\
FeaStNet-S~\cite{verma2018feastnet}, w/o refinement & XYZ            & 88.1\%            \\
FeaStNet-S~\cite{verma2018feastnet}, w/ refinement~\cite{vestner2017product}  & XYZ            & 92.2\%            \\ 
\hline
Ours, w/o refinement     & XYZ            & \textbf{94.0\%}   \\
Ours, w/ refinement~\cite{vestner2017product}     & XYZ            & \textbf{98.7\%}   \\ \hline\hline
\end{tabular}
\end{table}

\begin{figure*}[!h]
\centering
  \includegraphics[width=1.0\linewidth]{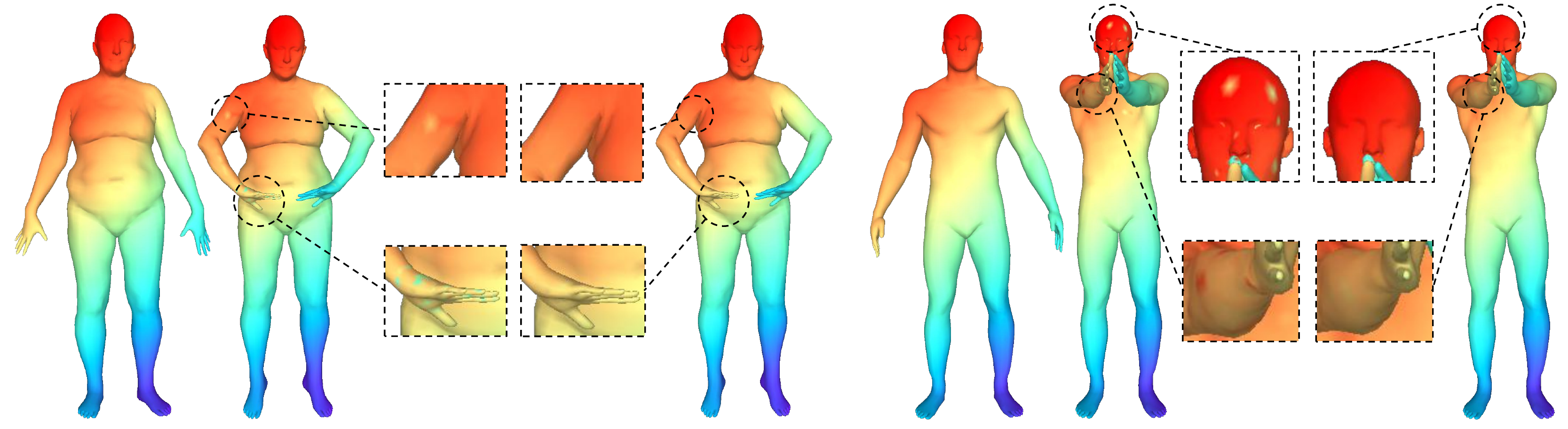}
  \caption{Two groups of correspondence results obtained by our approach on FAUST test set: reference shape (left), result without refinement (middle), result with refinement (right). Corresponding points are painted with same color.}
    \label{fig:faust_pred}
\end{figure*}

Figure~\ref{fig:faust_pred} visualizes correspondence results obtained by our model on the FAUST test set. Most correspondences are correctly matched without refinement, except for some areas with fine-grained local structures, such as areas on hands (the first example) and on the head (the second example). With refinement, these errors are rectified. Figure~\ref{fig:faust_error} visualizes correspondence errors in terms of the geodesic distance to the ground-truth correspondences. We notice that areas on/near joints, \eg, shoulder and waist, are prone to error, as these areas contain more detailed structural information that is hard to capture. With the help of refinement, such errors are rectified to some extent.

\subsection{Results on TOSCA}
\begin{figure}[!h]
\centering
  \includegraphics[width=1.0\linewidth]{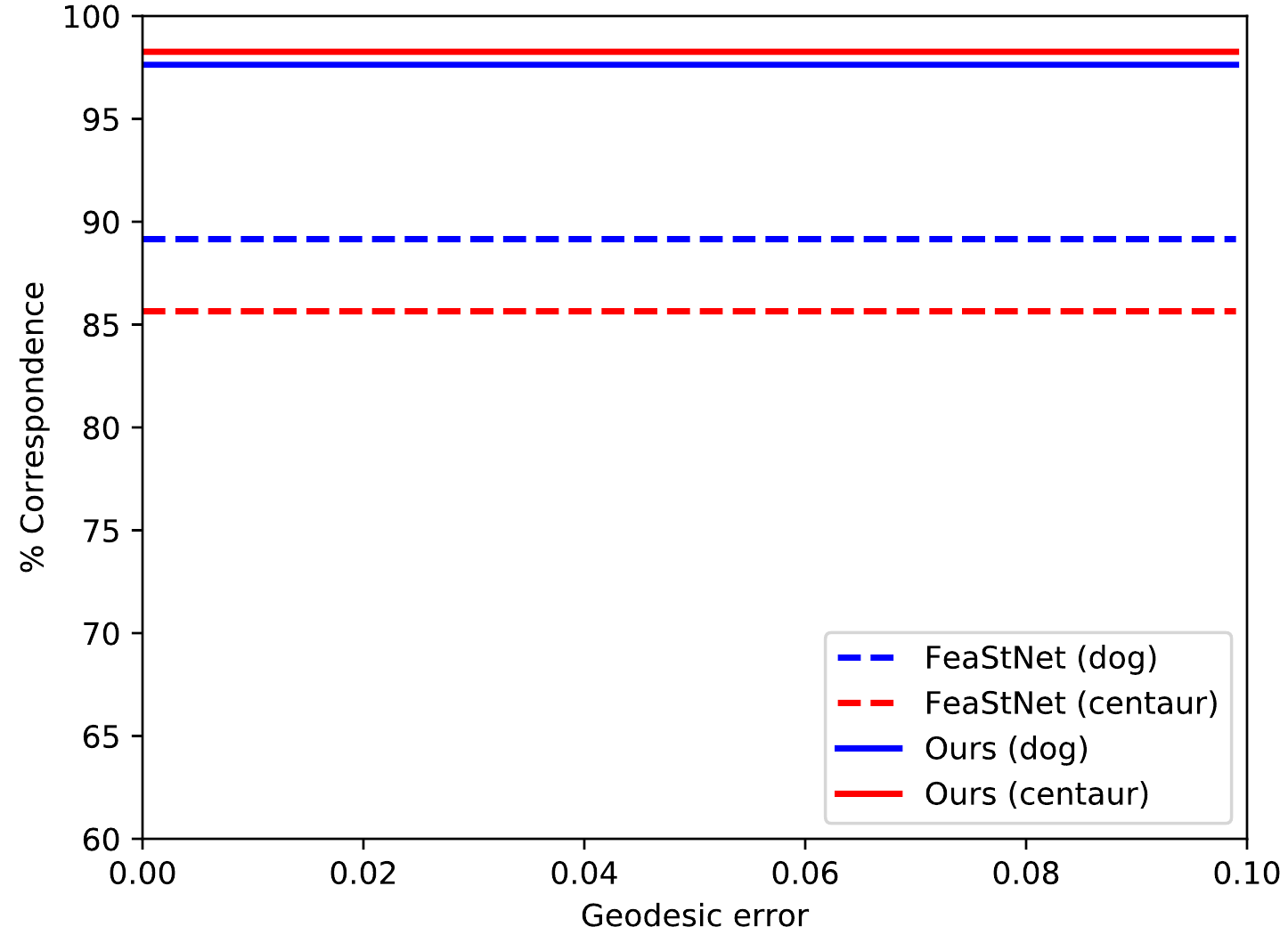}
  \caption{Comparison of fraction of geodesic shape correspondence errors within a certain distance with FeaStNet~\cite{verma2018feastnet} on dog and centaur categories. Note that they are trained on cat and horse categories separately.}
    \label{fig:tosca_distance}
\end{figure}

\begin{figure*}[!h]
\centering
  \includegraphics[width=1.0\linewidth]{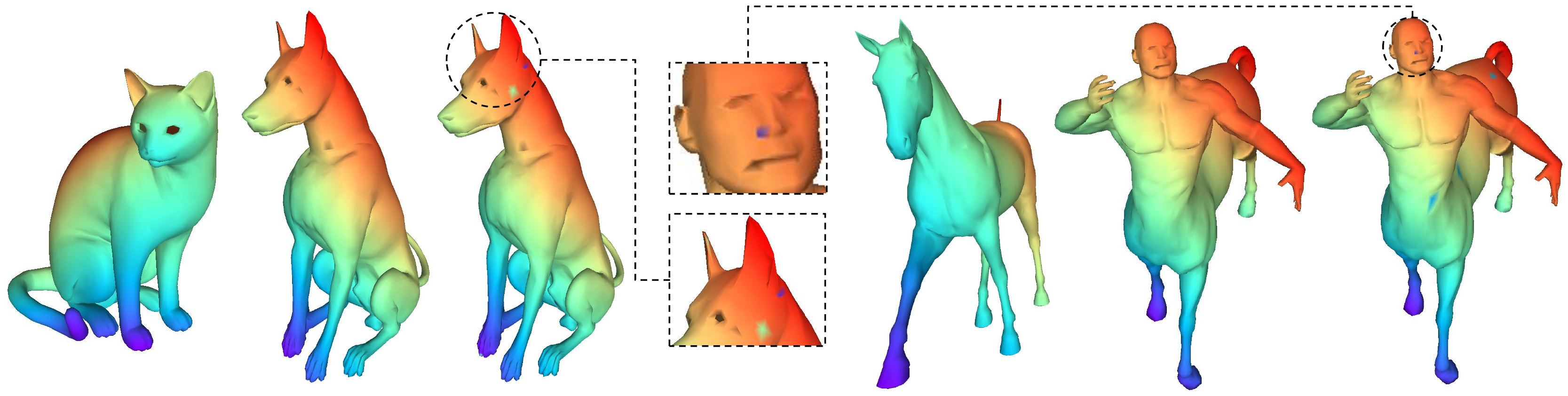}
  \caption{Two groups of correspondence results obtained by our approach on TOSCA dataset. For each group, training shape (left), ground-truth correspondence of testing shape (middle), predicted correspondence of testing shape (right). Corresponding points are painted with same color.}
    \label{fig:tosca_pred}
\end{figure*}

In Figure~\ref{fig:tosca_distance}, we plot the percentage of correspondences within a given geodesic distance from the ground-truth correspondence on the reference shape. We compare our method with FeaStNet~\cite{verma2018feastnet} under the same experimental settings for dog and centaur categories. As shown in the figure, our method outperforms FeaStNet by a large margin and achieves 97.5\% and 98.3\% correspondence accuracy with zero geodesic error for dog and centaur, respectively. Notice that as the training categories are different with the testing categories, \ie, trained on cat/horse but tested on dog/centaur, the results demonstrate the superiority of our method for shape correspondence on unseen categories. We attribute such generalization ability to the fact that the meta-learner learns the task distribution and then transfers local geometrical structure information across tasks from seen shapes to unseen shapes. Figure~\ref{fig:tosca_pred} visualizes correspondence results obtained by our model on TOSCA dataset. As shown in the figure, most correspondences in unseen shapes are correctly matched, except for some hard local areas that exhibit large discrepancy from the training shapes, such as areas on the nose of a human head.

\section{Analysis and Discussion}
In this section, we conducted further experiments to explore: 1. the robustness of our model against noise; 2. the distribution of local feature embeddings in the latent space; Here, we conduct all experiments on the FAUST test set without refinement post-processing.

\subsection{Resilience to Noise}
\begin{figure*}[!h]
  \centering
  \begin{minipage}[b]{0.52\linewidth}
    \includegraphics[width=1\linewidth]{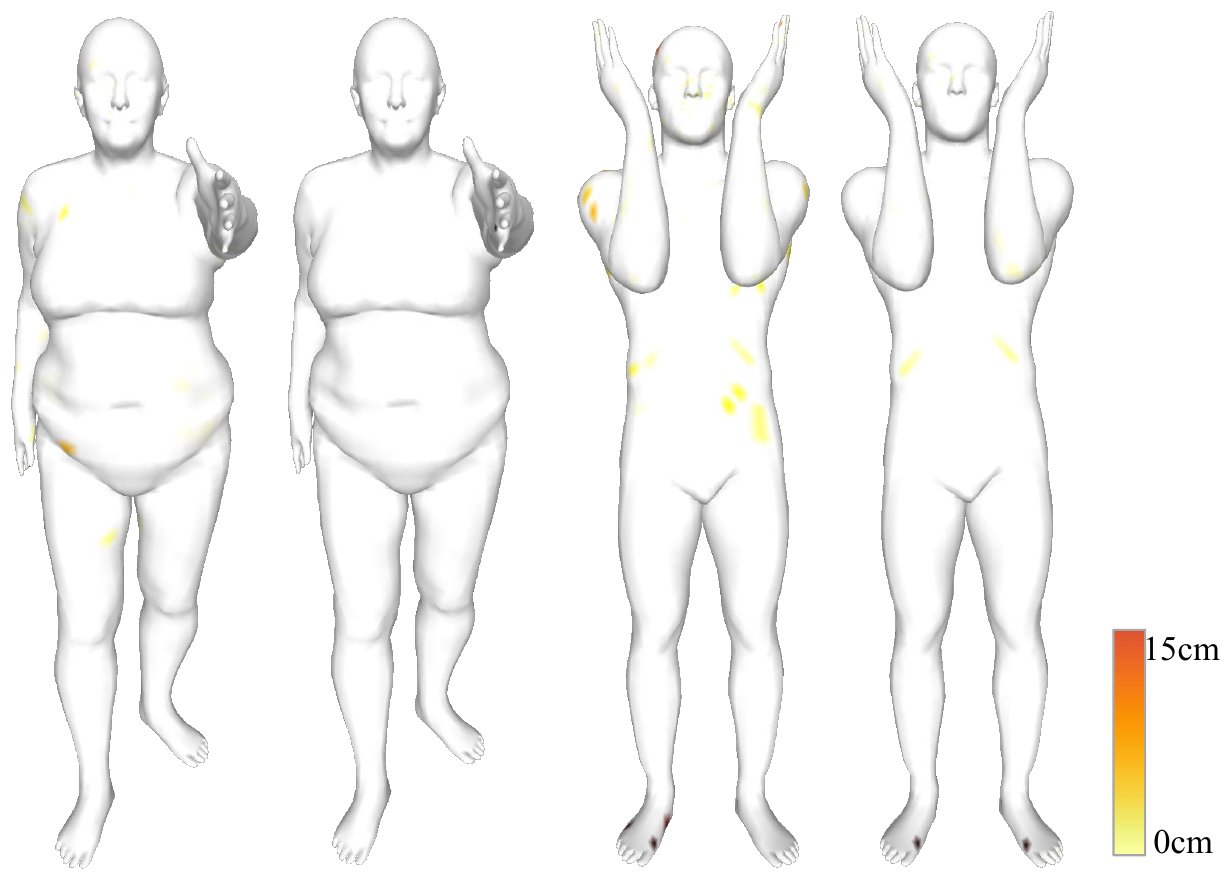}
    \caption{Geodesic errors on two FAUST test shapes: error without refinement (left), error with refinement (right).}
    \label{fig:faust_error}
  \end{minipage}
  \hfill
  \begin{minipage}[b]{0.46\linewidth}
    \includegraphics[width=1\linewidth]{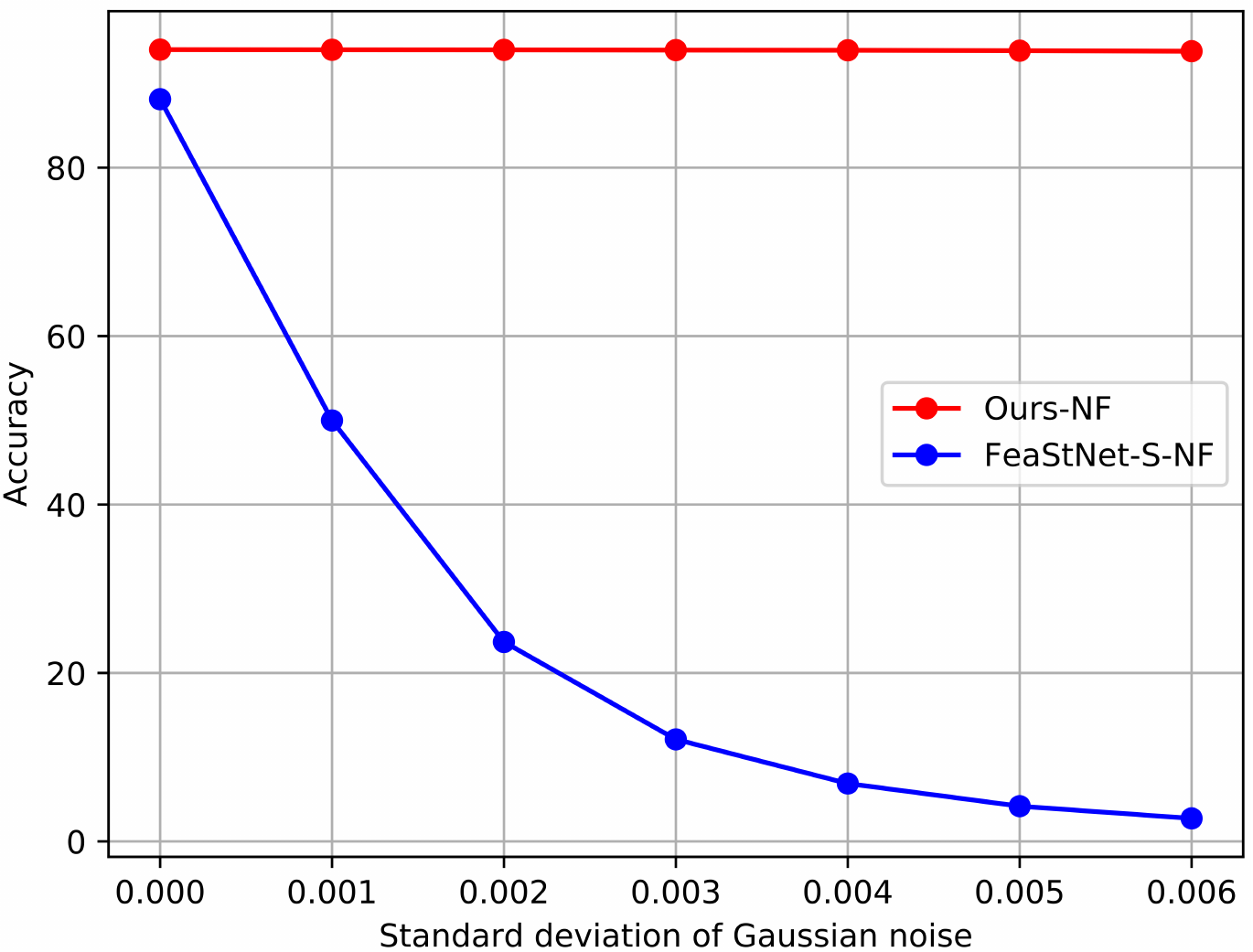}
\caption{Accuracy as a function of standard deviation of Gaussian noise added to FAUST test shapes.}
\label{fig:faust_noise}
  \end{minipage}
\end{figure*}

To evaluate our model's robustness, we add Gaussian noise to each vertex of the test shapes, and the distribution of noise is isotropic in three spatial (XYZ) dimensions. The mean is zero, and standard derivations are multi-scaled, ranging from 0.001 to 0.006 with equal intervals. We visualize the results of our model and the single-scale FeaStNet model in Figure~\ref{fig:faust_noise}. Note that both models are trained using noise-free data and tested on noisy data. From Figure~\ref{fig:faust_noise}, we can see that the FeaStNet model is much more noise sensitive than our model. The blue line dramatically drops with the increase of noise interference, while the red line almost keeps unchanged. It shows that the meta-learner in our model learns ``the process of learning 3D point signature'' and change base-learner's parameters adaptively, resulting in superior generalization ability on new tasks, \ie, unseen local geometric patterns.

\subsection{Clusters of Local Feature Embeddings}
\begin{figure*}[!h]
  \centering
    \includegraphics[width=1\linewidth]{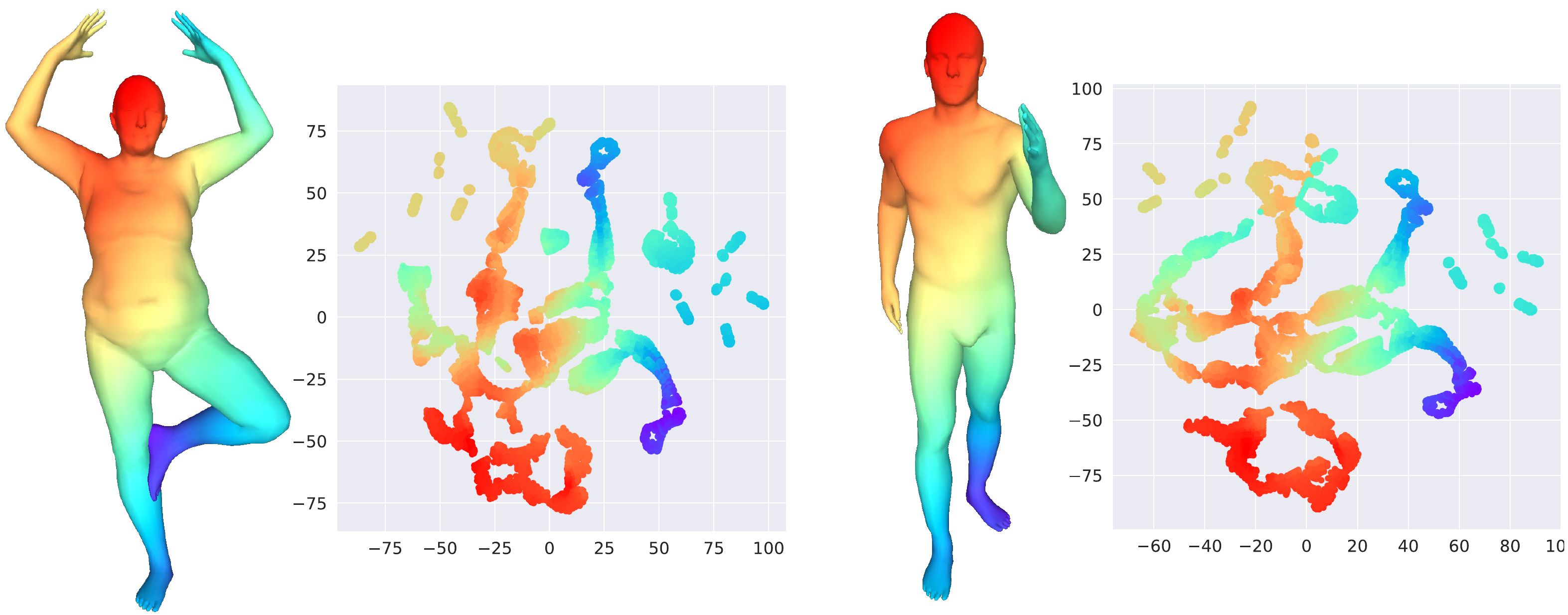}
    \caption{A t-SNE visualizations of local patch feature embeddings from shapes in FAUST test set.}
    \label{fig:faust_tsne}
\end{figure*}

In Figure~\ref{fig:faust_tsne}, we visualize local geometric feature embeddings obtained by the meta-learner network $g_\phi$ using t-SNE~\cite{maaten2008visualizing}. Specifically, each vertex corresponds to a local area in which the vertex is the center. Each area in 3D shapes and its corresponding projected two-dimensional embedding are plotted in the same color. From Figure~\ref{fig:faust_tsne}, we have two interesting observations: first, for a single shape, embeddings of local neighborhoods within the same body parts, \eg, head, arm, and feet, \etc, cluster together; second, across different shapes, embeddings of local neighborhoods within the same body parts locate roughly in the same regions in the 2D coordinates. Such observations indicate that $g_\phi$ captures local geometric structures and produces discriminative features. Therefore, weights of $f_\theta$ generated by $g_\phi$ based on neighborhood embeddings can model relations between pairs of neighboring vertices. Note that during the training process, we do not explicitly provide any embedding clustering signal for supervision, and therefore it demonstrates the inherent local structure modeling ability of our proposed method.

\section{Conclusion}
Learning point signature for 3D shapes using a meta-learning based method introduced in this work significantly improves the learned point signature quality over the baseline model. Our two-level networks, \ie, a base-learner and a meta-learner, can rapidly and reliably handle new signature generation tasks. Experimental results of 3D dense shape correspondence on two datasets achieve state-of-the-art performance, and demonstrate the ability of the proposed method to deal with unseen shapes. Furthermore, our method can be easily extended to different graph neural networks operating on irregular 3D geometric data. In the future, we plan to further develop our method to form a more general formulation such that it can be applied to more kinds of data modalities.


%





\ifCLASSOPTIONcaptionsoff
  \newpage
\fi



\bibliographystyle{IEEEtran}
\bibliography{IEEEabrv,ref}
\end{document}